\documentclass{article}
\usepackage{spconf,amsmath,graphicx,hyperref}
\usepackage{multirow}
\usepackage{booktabs}
\usepackage{amssymb}
\usepackage{enumitem}

\usepackage[table]{xcolor}

\definecolor{lightgray}{RGB}{242,242,242}

\usepackage{colortbl} 
\usepackage{xcolor}   
\definecolor{lightblue}{RGB}{230, 242, 255}

\title{ClearGCD: Mitigating Shortcut Learning for Robust Generalized Category Discovery}
\name{
Kailin Lyu$^{1,2}$,
Jianwei He$^{1}$,
Long Xiao$^{1,2}$,
Jianing Zeng$^{1}$,
Liang Fan$^{3}$,
Lin Shu$^{1}$,
Jie Hao$^{1,*}$
\thanks{*Corresponding author. Email: jie.hao@ia.ac.cn}
}
\address{
$^{1}$Institute of Automation, Chinese Academy of Sciences, Beijing, China \\
$^{2}$School of Artificial Intelligence, University of Chinese Academy of Sciences, Beijing, China \\
$^{3}$ Loughborough University, Loughborough, United Kingdom
}

\begin{document}
%
\maketitle
\begin{abstract}
In open-world scenarios, Generalized Category Discovery (GCD) requires identifying both known and novel categories in unlabeled data, but existing methods often face prototype confusion from shortcut learning, weakening generalization and causing forgetting of known classes. We propose \textbf{ClearGCD}, a framework that suppresses reliance on non-semantic cues through two complementary mechanisms: Semantic View Alignment (\textbf{SVA}), generating strong augmentations via cross-class patch replacement while enforcing semantic consistency with weak augmentations, and Shortcut Suppression Regularization (\textbf{SSR}), maintaining an adaptive prototype bank that aligns known classes and separates potential novel ones. ClearGCD can be seamlessly integrated into parametric GCD methods and consistently outperforms state-of-the-art methods across multiple benchmarks.
\end{abstract}
\begin{keywords}
semi-supervised learning, generalized category discovery, contrastive learning, shortcut learning
\end{keywords}
\section{Introduction}
\label{sec:intro}



Deep learning has made significant advances in vision tasks such as image recognition \cite{dl1,dl2,dl4}, but it relies on large labeled datasets and assumes unlabeled data share the same categories. This closed-world assumption breaks down in open-world settings with emerging unseen classes \cite{open1,open2}. To address this, Generalized Category Discovery (GCD) has been proposed to enable models to recognize both known and novel categories in unlabeled data, making it well-suited for large-scale open-world scenarios \cite{gcd,simgcd}.

Wen et al. \cite{simgcd} first proposed the parametric framework SimGCD, which replaces the computationally expensive clustering-based approach with a prototype classifier. Due to its competitive performance, it has become a robust baseline for GCD tasks. However, GCD still faces a central unresolved challenge: the confusion of representation prototypes \cite{protogcd}. When representations are biased, the prototypes of known and novel classes tend to overlap, resulting in degraded generalization on novel classes and catastrophic forgetting of known classes. To address this issue, LegoGCD \cite{legogcd} introduces additional regularization to alleviate forgetting of base classes, Ma et al. \cite{ma2025exploiting} aim to preserve semantic structure by optimizing sample selection and pseudo-label generation. In contrast to previous approaches that focus mainly on pseudo-label refinement and optimization of training strategy \cite{protogcd,legogcd,ma2025exploiting}, we approach the problem from the perspective of shortcut learning \cite{shortcut1,shortcut2}, revealing that the model’s reliance on background and contextual cues during representation learning is a key cause of prototype confusion.

\begin{figure}[!t]
    \centering 
    \includegraphics[width=\linewidth]{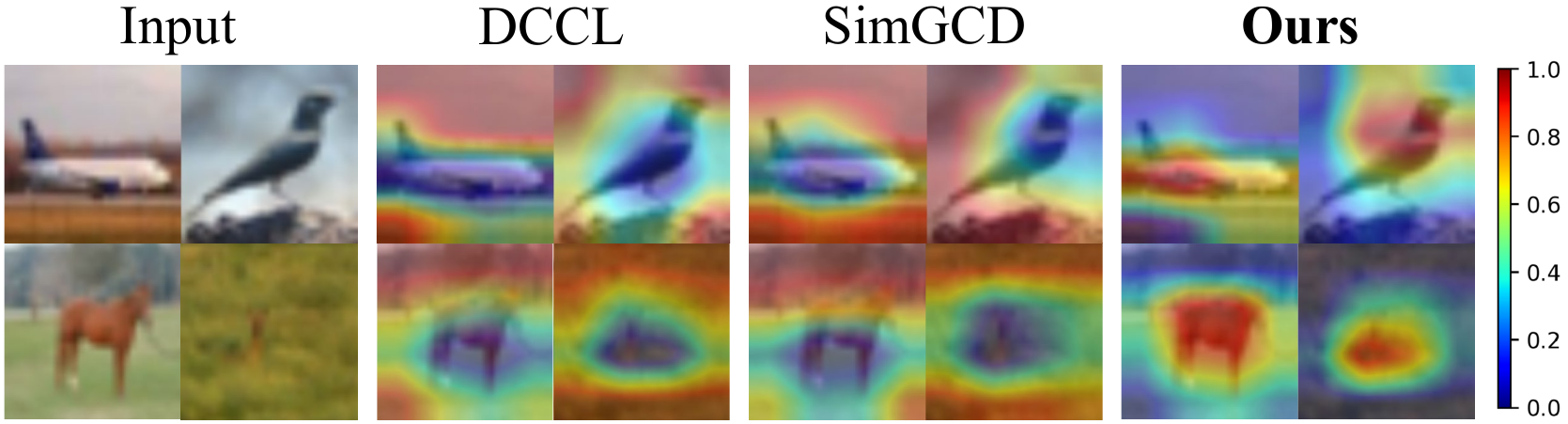} 
    \caption{\textbf{The visual explanations by GradCAM++ on the training set of CIFAR-10.} Although all methods predict the correct class, “shortcut learning” persists in DCCL \cite{dccl} and SimGCD \cite{simgcd}.}
    \label{fig:intro} 
    \vspace{-3mm}
\end{figure}

Intuitively, the neural network tends to “take shorcuts”  and focuses on simplistic features \cite{shortcut1,shortcut2}. This issue is particularly pronounced in GCD tasks, where models may learn biased features from labeled data and generate erroneous pseudo-labels, thereby compromising the classification of unlabeled samples. In practice, models often exploit scene-related information that is not related to target objects, which is a lack of generalization and is easily forgotten \cite{shortcut3,shortcut4}. For example, as shown in Figure 1, a model may rely on “blue sky” or “green grass” to distinguish between airplanes and horses. However, when novel categories such as birds or deer appear, similar backgrounds can invalidate previously learned knowledge and cause prototype confusion. Consequently, learning discriminative features that accurately represent categories is crucial to mitigating shortcut learning. Moreover, most existing GCD frameworks rely on self-distillation for knowledge transfer \cite{simgcd,protogcd}, whose effectiveness depends on the sufficiently discriminative features of old categories being preserved when learning new ones. However, due to shortcut learning, models often acquire oversimplified features that weaken generalization to novel categories, and self-distilling such biases further impairs novel class recognition.

To address the above issues, we propose ClearGCD, which aims to suppress shortcut learning and enhance the discriminability and robustness of representations. It consists of two key components: Semantic View Alignment (SVA) and Shortcut Suppression Regularization (SSR). SVA integrates strong augmentation by randomly replacing image patches across categories, while combining weak augmentation consistency constraints, directing the model toward semantic representations. Meanwhile, SSR imposes contrastive constraints that explicitly prevent novel samples from being misclassified due to scene similarity, while aligning known samples with their prototypes to alleviate catastrophic forgetting. The contributions are summarized as follows:



\begin{enumerate}\setlength{\itemsep}{0pt}\setlength{\parsep}{0pt}\setlength{\parskip}{0pt}
    \item We identify shortcut learning as a key limitation in GCD, in which biased features impair knowledge distillation and cause category confusion, thereby offering \textbf{new insights} for novel category recognition.
    \item We propose \textbf{ClearGCD}, integrating \textbf{SVA} and \textbf{SSR} to suppress shortcut learning and enhance representations, improving recognition in open-world settings.
    \item ClearGCD provides a \textbf{lightweight plug-and-play} mechanism that can be seamlessly integrated into parametric GCD and consistently outperforms state-of-the-art methods across multiple datasets.
\end{enumerate}


\section{METHOD}
\label{METHOD}


\subsection{Problem Formulation}
\label{Problem Formulation}

Generalized Category Discovery (GCD) aims to leverage knowledge from labeled data to adaptively cluster unlabeled data. We divide the entire dataset into two subsets: the labeled dataset \(\mathcal{D}_l = \{(x_i, y_i)\} \in \mathcal{X} \times \mathcal{Y}_l\), where \(\mathcal{Y}_l\) denotes the set of known categories, and the unlabeled dataset \(\mathcal{D}_u = \{(x_i, y_i)\} \in \mathcal{X} \times \mathcal{Y}_u\), where \(\mathcal{Y}_u\) spans all categories. Formally, we have \(\mathcal{Y}_l \subset \mathcal{Y}_u\). The goal of GCD is to categorize samples in \(\mathcal{D}_u\) by leveraging both the known category labels \(\mathcal{Y}_l\) and the unlabeled data itself. The total number of categories is denoted as \(K = |\mathcal{Y}_l \cup \mathcal{Y}_u|\), which is assumed to be known a priori, consistent with previous works \cite{simgcd,protogcd,legogcd}.

\subsection{Parametric clustering}
\label{Parametric clustering}

We adopt a parametric clustering framework similar to SimGCD \cite{simgcd}, consisting of two modules: \emph{representation learning} and \emph{parametric classification}. In the representation stage, we adopt a ViT-B/16 backbone \cite{vit} pretrained with DINO \cite{dino} on ImageNet, and perform supervised contrastive learning on labeled data and unsupervised contrastive learning on all data. Formally, given two augmentations $\tilde{x}_i$ and $x_i'$ of the same image in batch $B$, the latent features is $ z $, the unsupervised contrastive loss is
\begin{equation}
L_{\mathrm{rep}}^{u}=\tfrac{1}{|B|}\sum_{i\in B}-\log\frac{\exp(z_i^{\top}z_i'/\tau_u)}{\sum_{n\neq i}\exp(z_i^{\top}z_n'/\tau_u)},
\end{equation}
and the supervised contrastive loss, encouraging same-class features to cluster, is
\begin{equation}
L_{\mathrm{rep}}^{s}=\tfrac{1}{|B_l|}\sum_{i\in B_l}\tfrac{1}{|N_i|}\sum_{q\in N_i}-\log\frac{\exp(z_i^{\top}z_q'/\tau_c)}{\sum_{n\neq i}\exp(z_i^{\top}z_n'/\tau_c)}.
\end{equation}
The total representation loss is $L_{\mathrm{rep}}=(1-\lambda)L_{\mathrm{rep}}^{u}+\lambda L_{\mathrm{rep}}^{s}$.

For classification, we employs a self-distilled prototype classifier with $C=\{c_1,\ldots,c_K\}$, \(K = |\mathcal{Y}_l \cup \mathcal{Y}_u|\). For feature $h_i=f(x_i)$, $f$ is the feature backbone, the probability of class $k$ is
\begin{equation}
p_i^{(k)}=\frac{\exp\bigl(\tfrac{1}{\tau_s}(h_i/\|h_i\|_2)^{\top}(c_k/\|c_k\|_2)\bigr)}{\sum_{k'}\exp\bigl(\tfrac{1}{\tau_s}(h_i/\|h_i\|_2)^{\top}(c_{k'}/\|c_{k'}\|_2)\bigr)}.
\end{equation}
Using pseudo labels $q_i'$ for augmented views, the unsupervised and supervised classification losses are
\begin{equation}
L_{\mathrm{cls}}^{u}=\tfrac{1}{|B|}\sum_{i\in B}\ell(q_i',p_i),\quad L_{\mathrm{cls}}^{s}=\tfrac{1}{|B_l|}\sum_{i\in B_l}\ell(y_i,p_i).
\end{equation}
To stabilize unsupervised learning, we employ the mean-entropy regularizer from SimGCD \cite{simgcd}, defined as $H(p)=\sum_k p(k)\log p(k)$, where $p=\tfrac{1}{2|B|}\sum_{i\in B}(p_i+p_i')$. The classification objective is
\begin{equation}
L_{\mathrm{cls}}=(1-\lambda)(L_{\mathrm{cls}}^{u}-\epsilon H(p))+\lambda L_{\mathrm{cls}}^{s}.
\end{equation}

\begin{figure*}[htbp] 
    \centering 
    \includegraphics[width=0.8\textwidth]{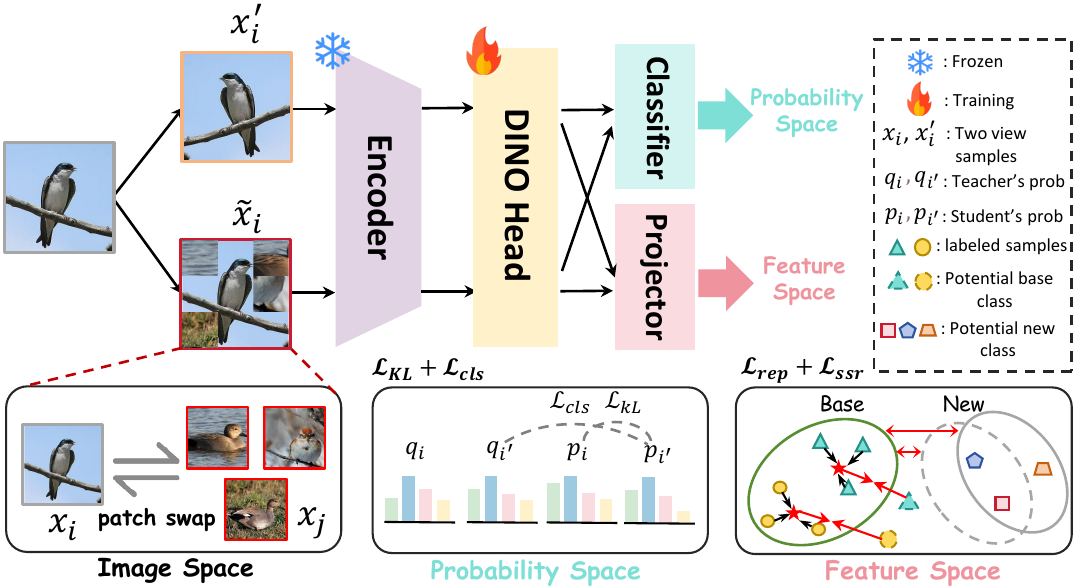}
    \caption{\textbf{The overview of our proposed framework in GCD.} By integrating SVA and SSR, we disrupt shortcut learning and enhance generalization performance in generalized category discovery.} 
    \label{fig:pipline}
    \vspace{-5mm}
\end{figure*}

\subsection{Semantic View Alignment}
\label{Semantic View Alignment}

As discussed in section~\ref{sec:intro}, shortcut learning often drives models to rely on spurious background cues rather than focusing on the true semantic content of objects, thereby aggravating prototype confusion. To address this issue, an intuitive strategy is to perform data augmentation at the image level to disrupt non-semantic correlations. Accordingly, we propose the \textbf{Semantic View Alignment (SVA)} module, which aims to reduce the model’s dependence on background or contextual information while enforcing semantic consistency across different augmentations. Specifically, SVA constructs strong augmentations by randomly selecting images from different categories within a mini-batch and performing patch replacement. Given a mini-batch 
\(\mathcal{B}=\{(x_i,y_i)\}_{i=1}^N\), for a sample \(x_i\), we randomly select another image \(x_j\) with \(y_i \neq y_j\) and generate a random binary mask \(\mathcal{M}\) (masking non-central image regions to avoid corrupting the primary subject). The augmented sample is then obtained as
\begin{equation}
\tilde{x}_i = \mathcal{M} \odot x_i + (1-\mathcal{M}) \odot x_j,
\end{equation}


where $\odot$ denotes element-wise multiplication and $M$ specifies the patch replacement between $x_i$ and $x_j$. The strongly augmented sample $\tilde{x}_i$ disrupts correlations with the original background, whereas a weakly augmented view $x_i'$ of the same image serves as a semantic anchor. We align the two representations with the following KL-based consistency loss:

\begin{equation}
\mathcal{L}_{\mathrm{KL}}
= \frac{1}{B/2}\sum_{i=1}^{B/2}
p(x_i)\,\log\frac{p(x_i')}{p(\tilde{x}_i)}\,,
\end{equation}
where $p(\cdot)$ is the predicted class-probability distribution.


\subsection{Shortcut Suppression Regularization}
\label{Shortcut Suppression Regularization}

Beyond preventing shortcut learning at the data augmentation level, we further impose constraints at the feature level to reduce the model’s reliance on background cues inherited from known categories when recognizing novel ones. To this end, we propose \textbf{Shortcut Suppression Regularization (SSR)}, which dynamically maintains class prototypes and applies contrastive constraints to suppress shortcut features. Specifically, for each known class $c \in \mathcal{C}_{known}$, we construct its prototype as $\mathbf{p}_c = \frac{1}{|D_l^c|} \sum_{x \in D_l^c} f(x)$, where $f(x)$ denotes the backbone representation and $\mathbf{p}_c$ is stored in a prototype bank that is periodically updated. For an unlabeled sample $x_u \in D_u$ with a pseudo-label predicted as a known class $c$, we apply a \textbf{positive alignment loss}:
\begin{equation}
\mathcal{L}_{PA}^{pos} = -\log \frac{\exp(\text{sim}(f(x_u),\mathbf{p}_c)/\tau)}{\sum_{k \in \mathcal{C}_{known}} \exp(\text{sim}(f(x_u),\mathbf{p}_k)/\tau)},
\end{equation}
where $\text{sim}(\cdot)$ denotes cosine similarity and $\tau$ is the temperature parameter. Conversely, if $x_u$ is predicted as a novel class, we introduce a \textbf{negative alignment loss}:
\begin{equation}
\mathcal{L}_{PA}^{neg} = -\frac{1}{|\mathcal{C}_{known}|} \sum_{c \in \mathcal{C}_{known}} \log \Big(1 - \sigma(\text{sim}(f(x_u),\mathbf{p}_c))\Big),
\end{equation}
where $\sigma(\cdot)$ is the sigmoid function. The overall SSR objective is defined as
\begin{equation}
\mathcal{L}_{SSR} = \mathcal{L}_{PA}^{pos} + \mathcal{L}_{PA}^{neg},
\end{equation}
which simultaneously enforces reliable alignment of samples with their semantic prototypes and separation of novel classes from known categories, thereby alleviating prototype confusion, preserving discriminability, and improving generalization to novel categories in open-world settings.



The overall loss function integrates the hierarchical contrastive learning loss with the classifier loss, incorporating both supervised and unsupervised components. Specifically, it is defined as:
\begin{equation}
L_{\text{total}} = \alpha \cdot (L_{\text{rep}} + L_{\text{cls}}+L_{\text{KL}}) + \beta \cdot L_{\text{SSR}},
\label{eq:total_loss}
\end{equation}
where $\alpha$ and $\beta$ are balancing parameters.

\section{EXPERIMENTS}
\label{EXPERIMENTS}


\begin{table*}[t]
\centering
\caption{Comparison of different methods on different datasets. The results of our method are averaged across three seeds.}
\resizebox{\textwidth}{!}{
\begin{tabular}{l|ccc|ccc|ccc|ccc|ccc|c}
\hline
\multirow{2}{*}{Methods} & \multicolumn{3}{c|}{CIFAR10 \cite{cifar10/100}} & \multicolumn{3}{c|}{CIFAR100 \cite{cifar10/100}} & \multicolumn{3}{c|}{ImageNet-100 \cite{imagenet100}} & \multicolumn{3}{c|}{CUB \cite{cub}} & \multicolumn{3}{c|}{FGVC-Aircraft \cite{aircraft}} & \multirow{2}{*}{Average} \\ \cline{2-16}
 & All & Old & New & All & Old & New & All & Old & New & All & Old & New & All & Old & New &  \\ \hline
$k$-means\cite{kmeans} & 83.6 & 85.7 & 82.5 & 52.0 & 52.2 & 50.8 & 72.7 & 75.5 & 71.3 & 34.3 & 38.9 & 32.1 & 16.0 & 14.4 & 16.8 & 51.7 \\
RS+\cite{RS} & 46.8 & 19.2 & 60.5 & 58.2 & 77.6 & 19.3 & 37.1 & 61.6 & 24.8 & 33.3 & 51.6 & 24.2 & 26.9 & 36.4 & 22.2 & 40.5 \\
UNO+\cite{uno} & 68.6 & \textbf{98.3} & 53.8 & 69.5 & 80.6 & 47.2 & 70.3 & \textbf{95.0} & 57.9 & 35.1 & 49.0 & 28.1 & 40.3 & 56.4 & 32.2 & 56.8 \\
ORCA\cite{orca} & 81.8 & 86.2 & 79.6 & 69.0 & 77.4 & 52.0 & 73.5 & 92.6 & 63.9 & 35.3 & 45.6 & 30.2 & 22.0 & 31.8 & 17.1 & 56.3 \\
PromptCAL\cite{PromptCAL} & \textbf{97.9} & 96.6 & 98.5 & 81.2 & \textbf{84.2} & 75.3 & 83.1 & 92.7 & 78.3 & 62.9 & 64.4 & \textbf{62.1} & 52.2 & 52.2 & 52.3 & 75.5 \\ 
GCD\cite{gcd} & 91.5 & 97.9 & 88.2 & 73.0 & 76.2 & 66.5 & 74.1 & 89.8 & 66.3 & 51.3 & 56.6 & 48.7 & 45.0 & 41.1 & 46.9 & 67.0 \\ \hline
SimGCD\cite{simgcd} & 97.1 & 95.1 & 98.1 & 80.1 & 81.2 & 77.8 & 83.0 & 93.1 & 77.9 & 60.3 & 65.6 & 57.7 & 54.2 & 59.1 & 51.8 & 74.9 \\
\rowcolor{lightblue}
\textbf{SimGCD+Ours} & 97.4 & 95.5 & 98.4 & 80.7 & 81.8 & 78.6 & 84.0 & 93.8 & 79.2 & 61.1 & 66.8 & 58.3 & 55.1 & 60.3 & 52.6 & 75.6\\ \hline
LegoGCD\cite{legogcd} & 97.1 & 94.3 & 98.5 & 81.8 & 81.4 & 82.5 & 86.3 & 94.5 & 82.1 & 63.8 & 71.9 & 59.8 & 55.0 & 61.5 & 51.7 & 76.8 \\
\rowcolor{lightblue}
\textbf{LegoGCD+Ours} & 97.7 & 95.8 & \textbf{98.6} & \textbf{82.4} & 82.3 & \textbf{82.7} & \textbf{87.3} & 94.8 & \textbf{83.6} & \textbf{65.5} & \textbf{73.7} & 61.4 & \textbf{55.7} & \textbf{62.5} & \textbf{52.3} & \textbf{77.7} \\ \hline
\end{tabular}
}
\label{tab:results}
\end{table*}

\begin{table}[ht]
\centering
\caption{Comparison of different methods on CUB and FGVC-Aircraft. Best results are in \textbf{bold}.}
\resizebox{\columnwidth}{!}{
\begin{tabular}{c| c c | c c c | c c c}
\toprule
\multirow{2}{*}{Schemes} & \multicolumn{2}{c|}{Component} & \multicolumn{3}{c|}{CUB} & \multicolumn{3}{c}{FGVC-Aircraft} \\
\cline{2-9}
 & SVA & SSR & All & Old & New & All & Old & New \\
\midrule
LegoGCD &  &  & 63.8 & 71.9 & 59.8 & 55.0 & 61.5 & 51.7 \\
a) & \checkmark &  & 65.0 & 72.2 & 61.5 & 55.1 & 61.3 & 52.1 \\
b) &  & \checkmark & 64.8 & 73.3 & 60.6 & 55.3 & 62.0 & 51.9 \\
\midrule
\rowcolor{lightgray}
\textbf{Ours} & \checkmark & \checkmark & \textbf{65.5} & \textbf{73.7} & \textbf{61.4} & \textbf{55.7} & \textbf{62.5} & \textbf{52.3} \\
\bottomrule
\end{tabular}
}
\label{tab:ablation}
\vspace{-3mm}
\end{table}

\subsection{Setup}
\label{Setup}

We systematically evaluated the effectiveness of the proposed method on five datasets, including general image recognition datasets (CIFAR10/100 and ImageNet-100) and fine-grained classification datasets (CUB and FGVC-Aircraft). For each dataset, we followed the experimental protocols of GCD and SimGCD by subsampling 50\% of the images from known classes in the training set to construct the labeled dataset $D^l$, while the remaining images from both known and novel classes formed the unlabeled dataset $D^u$. During training, we combined $D^l$ and $D^u$ into dataset $D$ for model learning. Furthermore, we strictly followed the parametric GCD \cite{gcd} evaluation protocol, retaining all the hyperparameter settings from SimGCD \cite{simgcd} and LegoGCD \cite{legogcd}, including the learning rate, batch size, temperature coefficient, and training epochs.


\subsection{Results and Discussion}
\label{Results and Discussion}

\begin{figure}[htbp]
    \centering 
    \includegraphics[width= \linewidth]{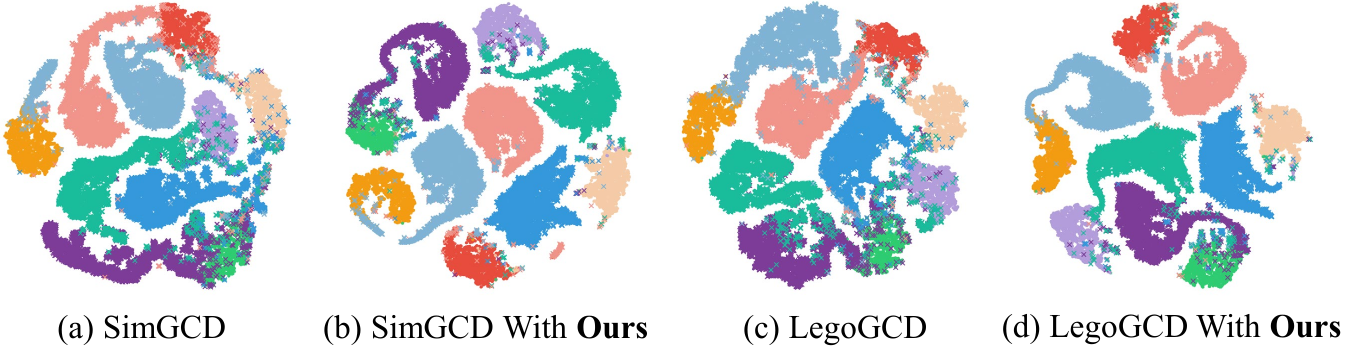} 
    \caption{Visualization of the representations of test data on CIFAR10 using t-SNE.}
    \label{fig:tsne} 
\end{figure}

We rigorously compared our method with multiple GCD baselines to further substantiate its superiority. Table~\ref{tab:results} presents the experimental results of all methods. Our method consistently improves SimGCD \cite{simgcd} and LegoGCD \cite{legogcd} across multiple benchmarks, with particularly strong performance in enhancing novel class recognition and mitigating catastrophic forgetting of old classes. We further visualize baseline feature representations with and without our method using t-SNE \cite{tsne}(Fig.~\ref{fig:tsne}). The results show markedly improved feature discriminability with our method. Notably, our approach can be seamlessly integrated into existing parametric GCD methods with minimal modifications, offering excellent plug-and-play generality. These performance gains primarily stem from the use of SVA and SSR, which effectively suppress shortcut learning and promote the acquisition of truly generalizable cross-instance features, thereby substantially enhancing the classifier’s ability to distinguish between categories.

\begin{figure}[ht]
    \centering 
    \includegraphics[width=\linewidth]{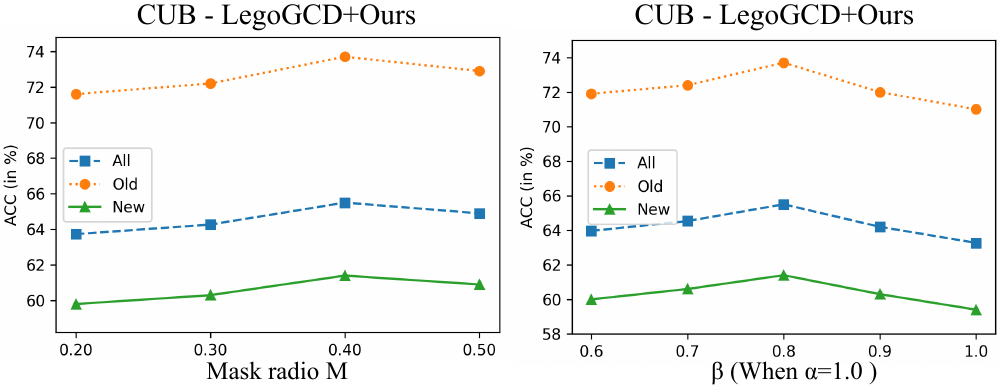} 
    \caption{Ablation study on $M$ and $\beta$ was conducted on CUB.}
    \label{fig:ablation} 
    \vspace{-3mm}
\end{figure}


In the ablation studies, We evaluated the contributions of different components to the GCD task (Table~\ref{tab:ablation}). Experiments demonstrate that, compared with LegoGCD \cite{legogcd}, incorporating these components improves performance on the “All,” “Old,” and “New” categories, highlighting the advantages of our design. Specifically, SVA effectively suppresses the coupling of models with irrelevant backgrounds at the data augmentation level, while SSR mitigates shortcut learning at the feature level to enhance inter-class discriminability, their complementary integration further boosts overall performance. In addition, we evaluated the impact of hyperparameters on performance (Fig.~\ref{fig:ablation}).

\section{Conclusion}
\label{Conclusion}

We propose ClearGCD, a lightweight plug-and-play framework that addresses prototype confusion and shortcut learning in GCD. By introducing Semantic View Alignment (SVA) and Shortcut Suppression Regularization (SSR), it suppresses reliance on non-semantic cues at both data augmentation and feature levels. Extensive experiments demonstrate that ClearGCD significantly improves novel class recognition and old class retention, consistently surpassing state-of-the-art methods for open-world category discovery.

\footnotesize
\bibliographystyle{IEEEbib}
\bibliography{strings,refs}

\end{document}